\documentclass[pdflatex,sn-mathphys-num]{sn-jnl}% Math and Physical Sciences Numbered Reference Style 
\usepackage{graphicx}%
\usepackage{subfig}
\usepackage{multirow}%
\usepackage{amsmath,amssymb,amsfonts}%
\usepackage{amsthm}%
\usepackage{mathrsfs}%
\usepackage[title]{appendix}%
\usepackage{xcolor}%
\usepackage{textcomp}%
\usepackage{manyfoot}%
\usepackage{booktabs}%
\usepackage{algorithm}%
\usepackage{algorithmicx}%
\usepackage{algpseudocode}%
\usepackage{listings}%
\usepackage{xcolor}  % Needed for \textcolor command

\newcommand{\eq}[1]{${#1}$}
\newcommand{\mycitename}[1]{{#1} \textit{et al.}}
\newcommand{\tablestyle}[2]{\setlength{\tabcolsep}{#1}\renewcommand{\arraystretch}{#2}\centering\footnotesize}
\theoremstyle{thmstyleone}%
\theoremstyle{thmstyletwo}%

\theoremstyle{thmstylethree}%

\begin{document}

\title[Article Title]{EVA-X: A foundation model for general chest X-ray analysis with self-supervised learning}

%%=============================================================%%
%% GivenName	-> \fnm{Joergen W.}
%% Particle	-> \spfx{van der} -> surname prefix
%% FamilyName	-> \sur{Ploeg}
%% Suffix	-> \sfx{IV}
%% \author*[1,2]{\fnm{Joergen W.} \spfx{van der} \sur{Ploeg} 
%%  \sfx{IV}}\email{iauthor@gmail.com}
%%=============================================================%%

\author[1]{\fnm{Jingfeng} \sur{Yao}}\email{jfyao@hust.edu.cn}

\author*[1]{\fnm{Xinggang} \sur{Wang}}\email{xgwang@hust.edu.cn}

\author[1]{\fnm{Yuehao} \sur{Song}}\email{yh\_song@hust.edu.cn}

\author[2]{\fnm{Huangxuan} \sur{Zhao}}\email{zhao\_huangxuan@sina.com}

\author[3,4,5]{\fnm{Jun} \sur{Ma}}

\author[1]{\fnm{Yajie} \sur{Chen}}\email{yajiechen@hust.edu.cn}

\author[1]{\fnm{Wenyu} \sur{Liu}}\email{liuwy@hust.edu.cn}

\author*[3,4,5,6,7]{\fnm{Bo} \sur{Wang}}\email{bowang@vectorinstitute.ai}

\affil[1]{\orgdiv{School of Electronic Information and Communications}, \orgname{Huazhong University of Science and Technology}, \orgaddress{\city{Wuhan}, \state{Hubei}, \country{China}}}

\affil[2]{\orgdiv{Department of Radiology, Union Hospital, Tongji Medical College}, \orgname{Huazhong University of Science and Technology}, \orgaddress{\city{Wuhan}, \state{Hubei}, \country{China}}}

\affil[3]{\orgdiv{Peter Munk Cardiac Centre}, \orgname{University Health Network}, \orgaddress{\city{Toronto}, \state{Ontario}, \country{Canada}}}

\affil[4]{\orgdiv{Department of Laboratory Medicine and Pathobiology}, \orgname{University of Toronto}, \orgaddress{\city{Toronto}, \state{Ontario}, \country{Canada}}}

\affil[5]{\orgname{Vector Institute for Artificial Intelligence}, \orgaddress{\city{Toronto}, \state{Ontario}, \country{Canada}}}

\affil[6]{\orgdiv{AI Hub}, \orgname{University Health Network}, \orgaddress{\city{Toronto}, \state{Ontario}, \country{Canada}}}

\affil[7]{\orgdiv{Department of Computer Science}, \orgname{University of Toronto}, \orgaddress{\city{Toronto}, \state{Ontario}, \country{Canada}}}

%%==================================%%
%% Sample for unstructured abstract %%
%%==================================%%

\abstract{
The diagnosis and treatment of chest diseases play a crucial role in maintaining human health. X-ray examination has become the most common clinical examination means due to its efficiency and cost-effectiveness. Artificial intelligence analysis methods for chest X-ray images are limited by insufficient annotation data and varying levels of annotation, resulting in weak generalization ability and difficulty in clinical dissemination. Here we present EVA-X, an innovative foundational model based on X-ray images with broad applicability to various chest disease detection tasks. EVA-X is the first X-ray image based self-supervised learning method capable of capturing both semantic and geometric information from unlabeled images for universal X-ray image representation. Through extensive experimentation, EVA-X has demonstrated exceptional performance in chest disease analysis and localization, becoming the first model capable of spanning over 20 different chest diseases and achieving leading results in over 11 different detection tasks in the medical field. Additionally, EVA-X significantly reduces the burden of data annotation in the medical AI field, showcasing strong potential in the domain of few-shot learning. The emergence of EVA-X will greatly propel the development and application of foundational medical models, bringing about revolutionary changes in future medical research and clinical practice. Our codes and models are available at \url{https://github.com/hustvl/EVA-X}.
}

\maketitle

\section{Introduction}

Chest X-rays constitute 40\% of the 3.6 billion imaging procedures performed annually worldwide due to their efficacy in diagnosing cardiopulmonary abnormalities, including COVID-19, pneumonia, pleural effusions, emphysema, and so on~\cite{world2016communicating, cid2024development}. This imaging technology provides several advantages, such as affordability, minimal radiation exposure, and widespread accessibility. The rapid evolution of Artificial Intelligence (AI) has led to the emergence of numerous deep learning models~\cite{cid2024development}, expediting the diagnostic process and improving the accuracy of X-ray image interpretation. However, these models encounter significant challenges. Their heavy reliance on extensive \textbf{labeled data}~\cite{chexpert, cxr14, mimic} not only consumes crucial medical resources but also limits their effectiveness and scalability in clinical settings. Moreover, the task-specific nature of current deep learning models restricts their ability to address diverse medical challenges, impacting their adaptability and flexibility in various healthcare settings.

AI foundational models have recently achieved outstanding milestones and become promising solutions to these challenges. Cutting-edge studies are rapidly expanding into medical research~\cite{moor2023foundation, zhou2023foundation, ma2024segment}. Trained on extensive datasets, these models provide precise diagnostic support, facilitating quicker and more accurate decisions for healthcare professionals. 
They are typically robust and versatile, achieving best performance across a wide range of healthcare scenarios. Their performance scalability is notable, increasing steadily with data and parameters to well adapt to diverse healthcare needs. Additionally, their interpretability enhances healthcare safety. These advantages eliminate the need for researchers to repeatedly and heavily annotate data and design specific deep-learning models for specific medical scenarios.
\textit{However, the medical domain has not yet seen an effective, flexible, scalable, and interpretable foundation model for chest X-ray images.}
% Previous efforts in X-ray detection have encountered challenges such as restricted training data, scalability limitations, and inadequate interpretability. There is an urgent need for a foundational model specifically for the detection of chest diseases.

\begin{figure}
    \centering
    \includegraphics[width=1\linewidth]{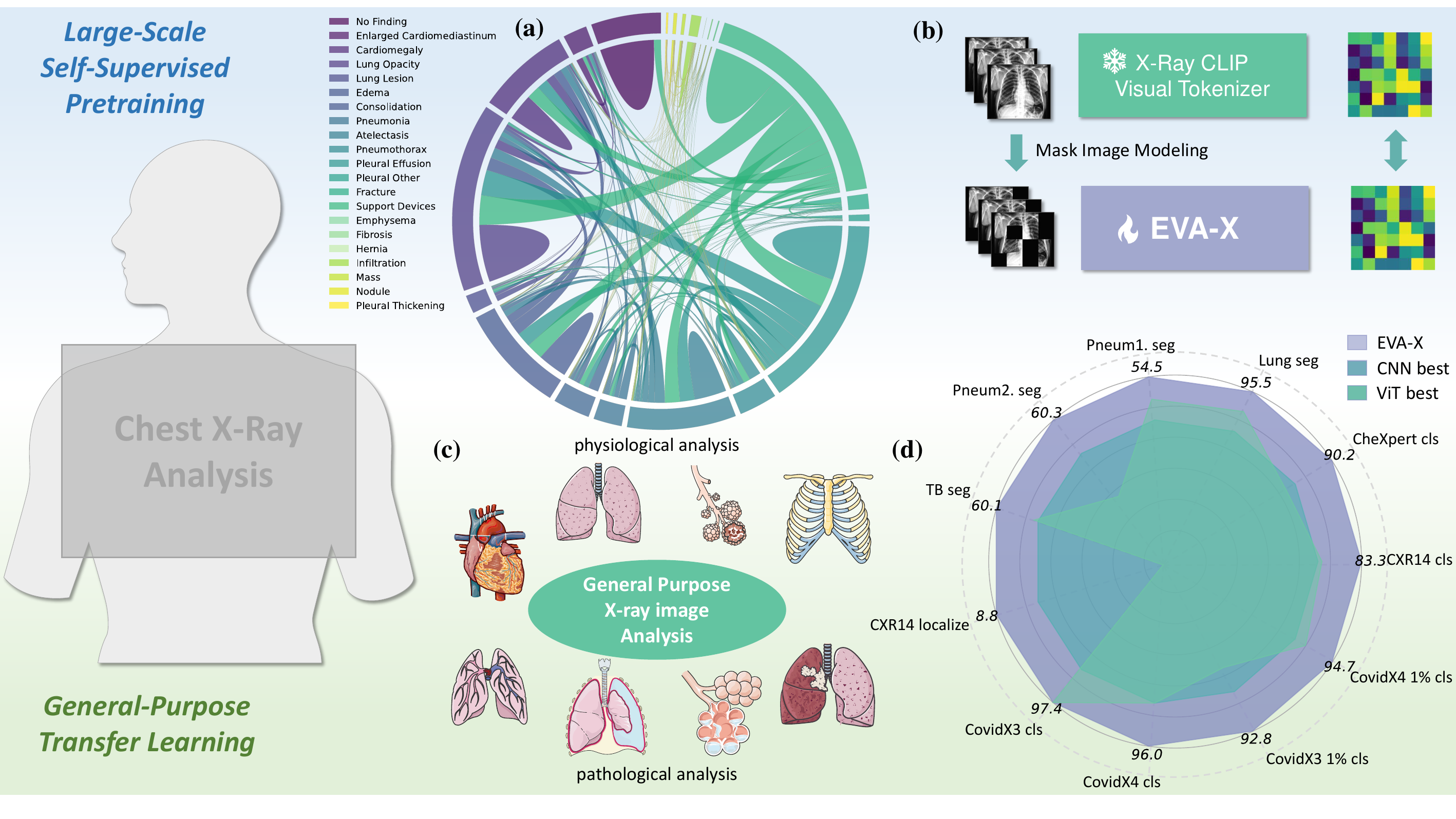}
    \caption{\textbf{(a) Pre-training Dataset}: EVA-X pre-training collects and leverages a diverse set of X-ray images encompassing various health conditions.~\cite{cxr14, chexpert, mimic} \textbf{(b) EVA-X Pre-training}: EVA-X employs a novel self-supervised pre-training approach that synergistically integrates the strengths of contrastive learning~\cite{mgca, biovil, medklip, convirt, gloria} and mask image modeling~\cite{selfmedmae, xiao2023delving}. \textbf{(c) General Visual Representations}: EVA-X exhibits a high degree of transferability, enhancing the comprehensive analysis of X-ray imagery. \textbf{(d) Transfer Performance}: EVA-X demonstrates state-of-the-art performance across 11 distinct tasks~\cite{cxr14, chexpert, hemdan2020covidx, zhang2023lung, siim, RSNA, jaeger2014two}, outperforming established benchmarks set by previous pre-trained models.}
    \label{fig:first-fig}
\end{figure}

In response, we introduce EVA-X, a foundational model for comprehensive chest X-ray analysis using self-supervised learning. Leveraging extensive \textbf{unlabeled} data, EVA-X acquires \textbf{general} visual representations, enabling effective deployment across all chest disease detection tasks based on X-rays.
EVA-X demonstrates significant technological advancement by not requiring annotated data for training, thus reducing the demand for medical resources compared to traditional contrastive learning methods~\cite{mgca, biovil, medklip, convirt, gloria}. Moreover, EVA-X pioneers a strategy in the X-ray domain to simultaneously learn semantic and geometric features, combining the advantages of contrastive learning pre-training~\cite{mgca, biovil, medklip, convirt, gloria} and mask image modeling pre-training~\cite{selfmedmae, xiao2023delving}. This innovative approach enhances the universality of its visual representations, facilitating broad utilization across diverse chest disease detection tasks and showcasing exceptional generalization capabilities. 

Extensive experiments have demonstrated the superiority of EVA-X in the X-ray domain. From the perspective of pre-trained visual representations, EVA-X is capable of learning without using any annotated data. Compared to 15 previous pre-trained models~\cite{vit, resnet, densenet, medklip, mgca, biovil, medklip, selfmedmae, xiao2023delving, eva02}, EVA-X exhibits greater scalability and flexibility. From the standpoint of transfer learning, we tested EVA-X on 11 X-ray physiological and pathological analysis tasks. The results indicate that EVA-X has significant advantages in semantic understanding and geometric analysis. Moreover, EVA-X can significantly reduce the need for annotated data in downstream tasks. For instance, in COVID-19 detection, EVA-X achieves a 95\% accuracy with just 1\% of the training data. In terms of interpretability, EVA-X can determine lesion locations using only category information. We argue that EVA-X holds the potential to significantly enhance AI's diagnostic performance in chest diseases, thereby broadening the application scope of AI within healthcare, reducing the strain on medical resources, and ultimately contributing to the promotion of global public health.

\begin{figure}
    \centering
    \includegraphics[width=\linewidth]{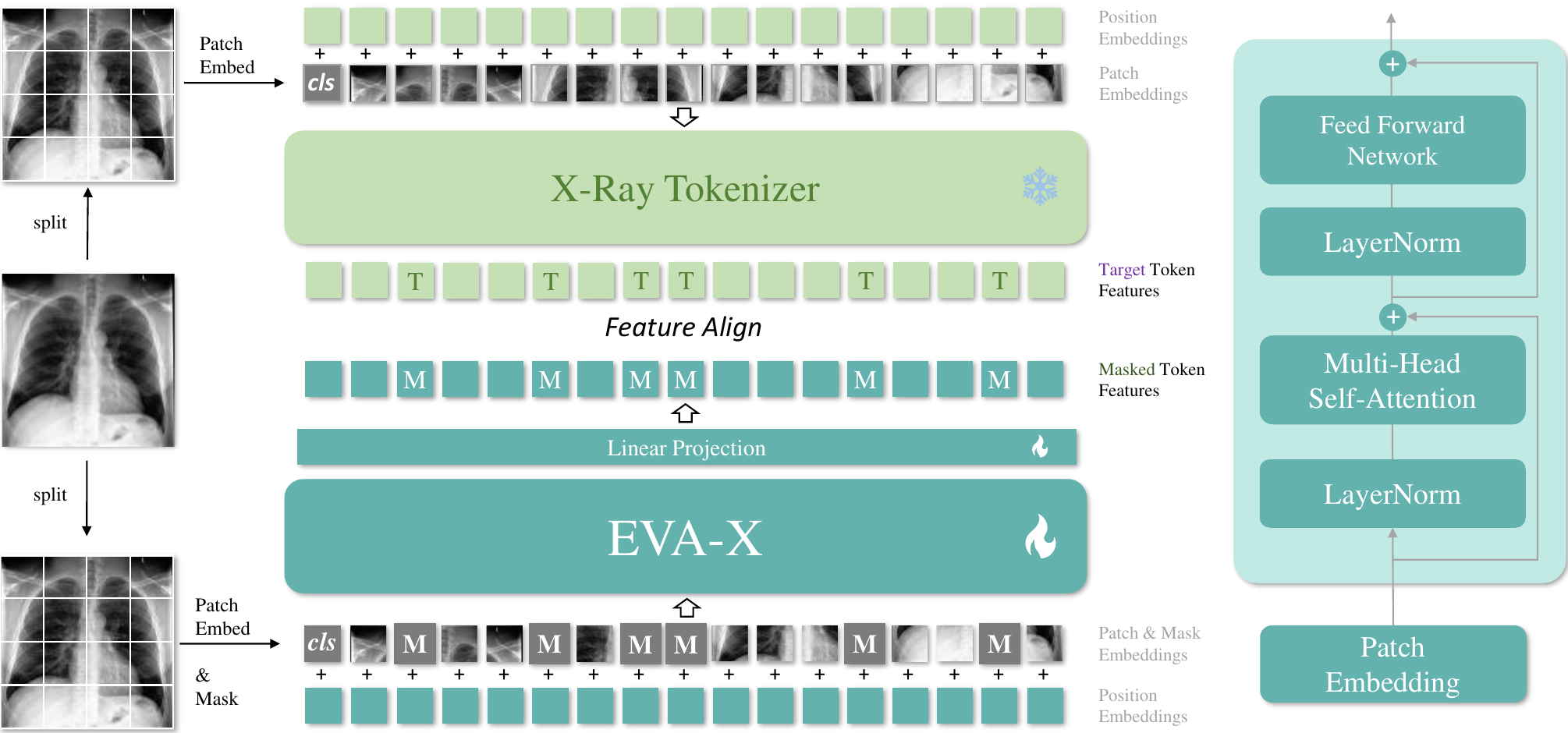}
    \caption{\textbf{Overall of EVA-X self-supervised pre-training.} EVA-X designs a self-supervised pre-training method combining the advantages of contrastive learning and mask image modeling for Chest X-ray images. Please see Sec.~\ref{sec:method} for details.}
    \label{fig:enter-label}
\end{figure}

\section{Results}\label{sec2}

% Something about Figure1
EVA-X is a family of medical foundational models pre-trained specifically for analyzing and diagnosing chest diseases. It utilizes the widely adopted vision transformer architecture~\cite{vit} in computer vision and acquires general visual representations through unlabeled X-ray images.

Illustrated in Fig~\ref{fig:first-fig} (a), our pre-training dataset encompasses more than 20 distinct human chest health conditions, reflecting the diversity and complexity of chest health issues. EVA-X designs a novel self-supervised pre-training approach for X-ray images (Fig~\ref{fig:first-fig} (b)). This approach combines the benefits of contrastive learning and mask image modeling, efficiently capturing semantic and geometric information without requiring manual annotations during training. Due to its diverse training data and superior self-supervised training design, EVA-X can generalize to various X-ray-based chest disease detection scenarios. It is applicable to a wide range of tasks in chest physiology and pathology analysis (Fig~\ref{fig:first-fig} (c)). We evaluate EVA-X's performance on 11 different X-ray image analysis tasks and compare it with the previous best methods. As depicted in Fig~\ref{fig:first-fig} (d), EVA-X outperforms all of them, achieving state-of-the-art (SoTA) results across all tasks. To our knowledge, EVA-X represents a comprehensive advancement of the advanced ViT structure over traditional convolutional models in the medical X-ray domain. This innovation heralds a new era in X-ray technology, where robust visual foundational models are likely poised to replace traditional designs.

Below, we analyze the superiority of EVA-X in detail from three major perspectives: pre-training, transfer learning, and interpretability. We discuss the EVA-X self-supervised learning method in Section~\ref{sec:method}, as illustrated in Figure~\ref{fig:enter-label}.

% The unique training system of EVA-X, which learns without the need for manual annotation, enables us to effectively aggregate and utilize all publicly available large-scale X-ray image databases. In contrast, traditional methods dependent on consistent annotation standards are severely restricted, and typically effective only within a single dataset. As illustrated in Fig~\ref{fig:first-fig} (b), our pre-training dataset contains over 20 different human chest health conditions, reflecting the diversity and complexity of chest health issues. EVA-X demonstrates strong scalability, with its performance significantly enhanced as the training data and parameter size increase. This characteristic allows it to better adapt to application scenarios under different computational resource conditions. Due to this diversity of training data and the model's flexibility, EVA-X can be generalized to all X-ray-based chest disease detection scenarios. Additionally, EVA-X also demonstrates strong few-shot learning capabilities, achieving a 95\% detection accuracy on a COVID-19 dataset with just 1\% of labeled training data (about 300 samples). Extensive experiments have demonstrated that EVA-X surpasses the previous best methods in nine tasks, including chest disease detection, COVID-19 detection, chest organ segmentation, and chest disease lesion segmentation, achieving new state-of-the-art performance (Fig~\ref{fig:first-fig} (e)). 

\subsection{Pre-training: Performance, Efficiency and Flexibility}
\label{sec:pre-training}

We evaluate the EVA-X pre-training method across three dimensions: the performance of pre-trained visual representations, the number of parameters, and computational FLOPs. Our evaluation employs the CXR14 test set~\cite{cxr14}, which serves as the benchmark dataset in the X-ray domain (see Sec~\ref{sec:method-dataset}). We compare EVA-X with 15 different pre-trained X-ray models, including widely used models such as DenseNet121~\cite{densenet}, ResNet50~\cite{resnet}, and ViTs~\cite{vit}. Considering the diverse computational demands of medical scenarios, we train three EVA-X models of different scales: EVA-X-Ti, EVA-X-S, and EVA-X-B.

\paragraph{Performance}

\begin{figure}
    \centering
    \includegraphics[width=1\linewidth]{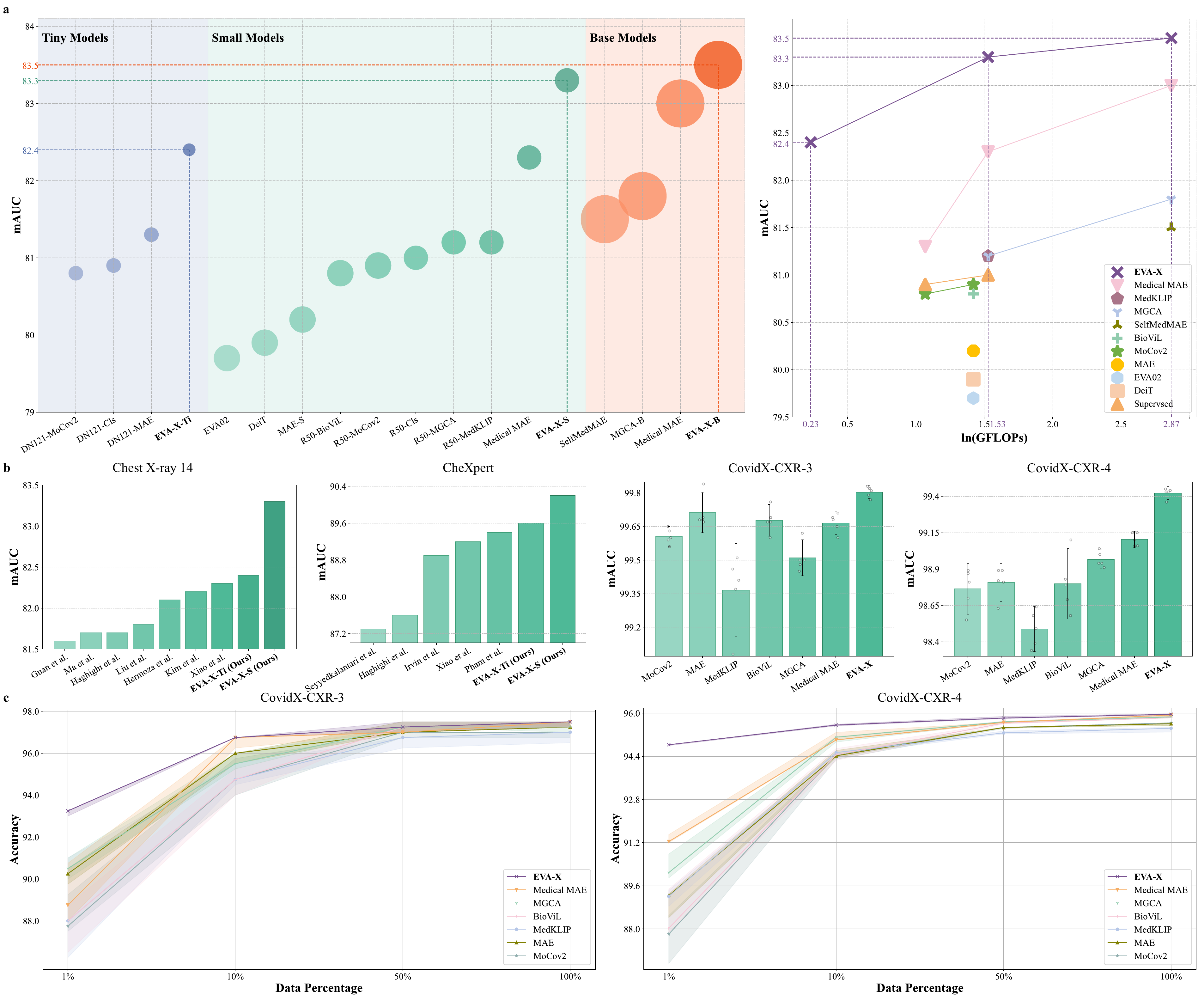}
    \caption{\textbf{(a) Performance and Efficiency of EVA-X Pre-trained Models.} Among all pre-trained models~\cite{vit, resnet, densenet, medklip, mgca, biovil, medklip, selfmedmae, xiao2023delving, eva02}, EVA-X-B achieves the highest performance. The EVA-X family demonstrates an excellent balance between performance and computational efficiency compared to previous methods. \textbf{(b) Performance on Chest Diseases Classification.} EVA-X achieves the best performance in both multi-label and single-label classification tasks for chest diseases~\cite{cxr14, chexpert, hemdan2020covidx}. \textbf{(c) Performance on Label-efficient Classification.} EVA-X shows superior performance across varying amounts of training data, with a particularly notable advantage observed when dealing with extremely limited training data.}
    \label{fig:result1}
\end{figure}

As depicted in Fig~\ref{fig:result1} (a) left, we categorize these 18 different pre-trained models~\cite{vit, resnet, densenet, medklip, mgca, biovil, medklip, selfmedmae, xiao2023delving, eva02} into three comparison groups: tiny models, small models, and base models, based on their parameter counts. Notably, within each group, EVA-X consistently exhibits the lowest parameter count (6M, 22M, 86M). We observe remarkable scalability in EVA-X, with its performance consistently improving as the parameter count increases. Among these models, EVA-X-B stands out as the best pre-trained X-ray model, achieving a visual representation test performance of 83.5 mAUC, surpassing all previous medical self-supervised pre-training methods such as Medical MAE~\cite{xiao2023delving}, contrastive learning pre-training methods like MGCA~\cite{mgca}, and well-known pre-training methods for natural images like MAE~\cite{mae} and MoCov2~\cite{mocov2}. This achievement sets a new standard for state-of-the-art performance in medical X-ray pre-training.

\paragraph{Efficiency}

As depicted in Fig~\ref{fig:result1} (a) right, we assess the computational complexity of all methods during testing. To facilitate visualization, we logarithmically scale the FLOPs on the horizontal axis. The purple \textcolor{purple}{x} marker on the graph signifies the correlation curve between computational complexity and the performance of EVA-X. EVA-X strikes an outstanding balance between performance and computational complexity compared to all other methods.

\paragraph{Flexibility}
Typically, foundational models aiming for high performance often impose high computational demands and it could be challenging in resource-constrained medical environments. 
However, leveraging the impressive capabilities of EVA-X, we not only investigate its performance boundaries but also develop a lightweight variant, EVA-X-Ti. It is worth noting that EVA-X-Ti is the model with the lowest computational complexity (1.26 GFLOPs) among them with incredible performance (82.4 mAUC). We conduct comparative experiments between EVA-X-Ti and 15 previously introduced pre-trained models, most of which have larger parameter counts than EVA-X-Ti. Despite this, EVA-X-Ti, with its streamlined parameters (6M), outperformed 14 of these models in performance metrics. It even outperforms MGCA-B~\cite{mgca} (81.8 mAUC) and SelfMedMAE~\cite{selfmedmae} (81.5 mAUC), which have 13 times more FLOPs than EVA-X-Ti. This exceptional performance highlights EVA-X-Ti's potential as a cost-effective alternative to large-scale models, promoting wider adoption and deeper integration of EVA-X technology across various applications.

\begin{figure}
    \centering
    \includegraphics[width=\linewidth]{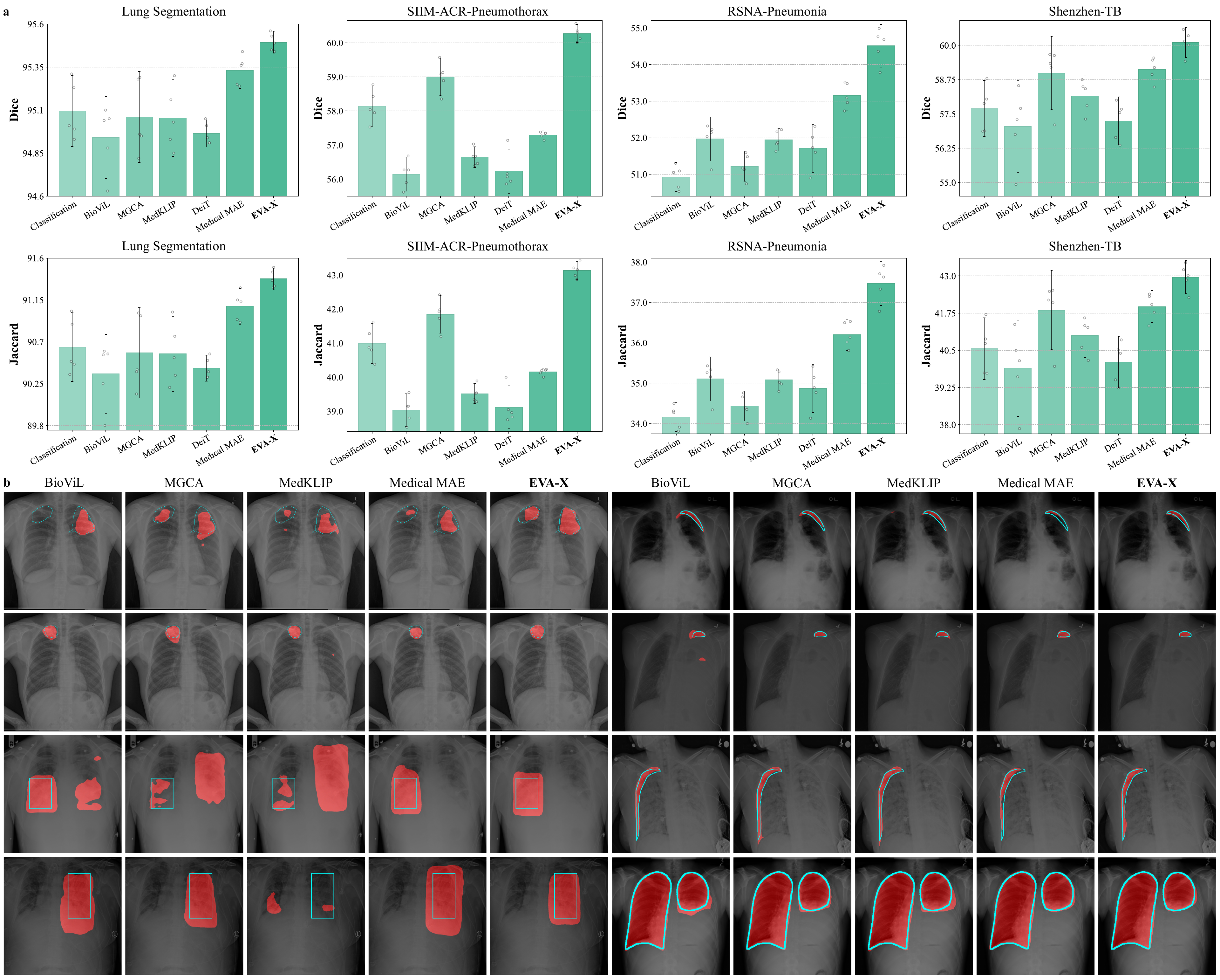}
    \caption{\textbf{(a) Performance on Chest X-ray Segmentation.} EVA-X surpasses 6 other pre-trained models~\cite{resnet, medklip, mgca, deit, biovil, xiao2023delving} across all segmentation benchmarks~\cite{zhang2023lung, RSNA, siim, jaeger2014two}, exhibiting superior performance on Dice and Jaccard metrics. \textbf{(b) Visualization of Segmentation Results.} EVA-X demonstrates enhanced accuracy and finer masks across all segmentation tasks.}
    \label{fig:result2}
\end{figure}

\begin{figure}
    \centering
    \includegraphics[width=0.9\linewidth]{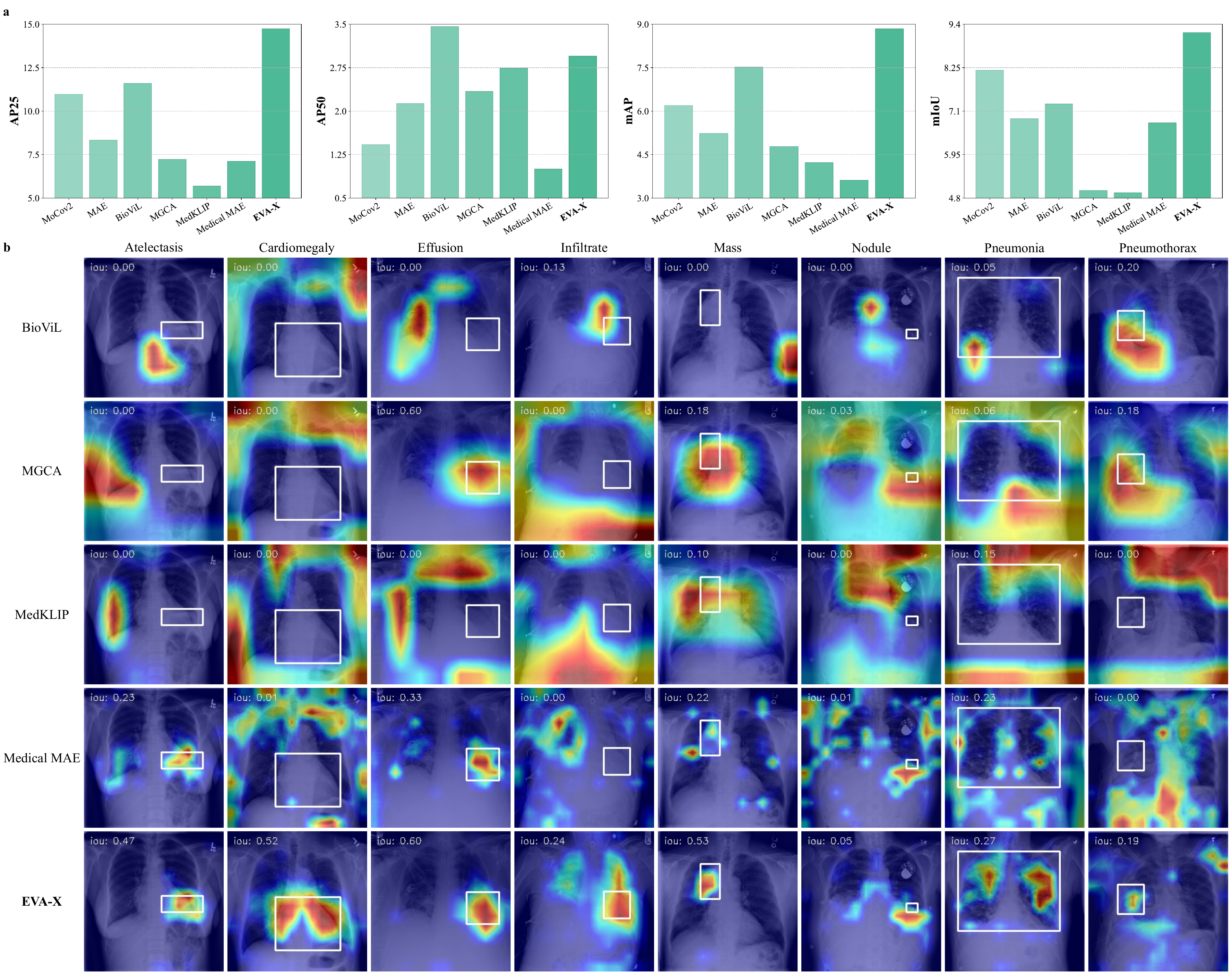}
    \caption{\textbf{(a) Performance on Weakly-Supervised Localization.} EVA-X delivers the highest overall performance across all four metrics for weakly supervised localization tasks.~\cite{cxr14} \textbf{(b) Visualization of Grad-CAM.} Class Activation Map (CAM)~\cite{gradcam} is a significant interpretation method for deep learning models. It illustrates that EVA-X can localize diseases using only classification annotations, showcasing remarkable interpretability.}
    \label{fig:result3}
\end{figure}

\subsection{Transfer Learning to X-ray image analysis}

\paragraph{Chest Diseases Classification}

X-ray images are one of the important tools for diagnosing chest diseases, with different diseases exhibiting different manifestations on X-ray images. Our experiments demonstrate that the visual representations learned by EVA-X pre-training are universal and can be generalized to diagnostic tasks for all chest diseases.

\textbf{Multi-label classification} requires the model to make judgments about the presence of multiple different diseases at once.
In our work, we evaluate the general disease detection capability of EVA-X using two commonly used multi-class chest disease diagnosis datasets, Chest X-Ray14~\cite{cxr14}, and CheXpert~\cite{chexpert}. We fine-tune the visual representations learned by EVA-X on these two datasets without employing any additional design techniques. 

As shown in Figure~\ref{fig:result1} (b) CXR14, we compare the results of EVA-X with 8 different methods~\cite{guan2020multi, ma2019multi, haghighi2022dira, liu2022acpl, hermoza2020region, kim2021xprotonet, xiao2023delving} on the Chest X-Ray14 dataset. Most of these methods are designed for chest X-ray classification. Among them, our EVA-X-Ti (6M) with 82.4 mAUC exceeds the 82.2 mAUC achieved by Kim et al~\cite{kim2021xprotonet}. Their method uses DenseNet121 (8M) as a backbone. Our EVA-X-S (22M) with 83.3 mAUC, exceeds the 82.3 mAUC achieved by Xiao et al.~\cite{xiao2023delving} with ViT-S 0.823 mAUC. Taken together, EVA-X outperforms the previous best method at two different sizes, reaching new SOTA results. From the perspective of single-disease diagnosis, EVA-X performs best by achieving the highest accuracy in 12 out of 14 disease diagnoses (see appendix for more details). 

As shown in Figure~\ref{fig:result1} (b) CheXpert, we compare EVA-X with 5 previous methods~\cite{seyyed2020chexclusion, haghighi2022dira, chexpert, xiao2023delving, pham2021interpreting}. In terms of individual metrics, EVA-X reaches new SOTA results in 2 categories (see appendix for more details). In terms of mAUC, both EVA-X-Ti, and EVA-X-S outperform all previous methods and reach new SOTA results. Among them, EVA-X-Ti has only 6M parameters, which is smaller than all previous methods exceeds the performance of all previous methods, and achieves new SOTA results.

\textbf{Single-label classificiation} requires the model to make accurate judgments about a specific disease. In this paper, we test this using COVID-19 as an example. Specifically, we utilize the latest collected and annotated datasets COVID-CXR-3 and COVID-CXR-4~\cite{hemdan2020covidx} and fine-tune 7 different pre-trained models~\cite{resnet, medklip, mgca, deit, biovil, xiao2023delving}, including EVA-X, on each dataset. 
As shown in Figure~\ref{fig:result1} (b) CovidX-CXR-3 and CovidX-CXR-4, EVA-X ranks first among all methods with exceptionally high 99.8 mAUC and 99.4 mAUC. Additionally, EVA-X maintains remarkable stability, demonstrating the most consistent performance across multiple experiments. Specifically, the mean standard deviation of EVA-X on both datasets is 0.03, which is lower than all other methods including Medical MAE~\cite{xiao2023delving} (0.045), MGCA~\cite{mgca} (0.055), BioViL~\cite{biovil} (0.135), etc.

\paragraph{Label Efficient Classification}

The EVA-X model, optimized through large-scale data pre-training, exhibits a high sensitivity to small training data in downstream tasks. It can converge rapidly with minimal data, thereby directly alleviating the pressure of annotation data on the healthcare system. In Fig~\ref{fig:result1} (c), we validate EVA-X's efficient training capability on COVID-19~\cite{hemdan2020covidx} and compare it with previous methods~\cite{xiao2023delving, mgca, biovil, medklip, mae, mocov2}.
EVA-X demonstrates the strongest and most stable performance at different data sizes. Especially in the case of very little annotated data, only 1\% training data, EVA-X shows a clear advantage over other methods. On the CovidX-CXR-4 dataset, EVA-X achieves 95\% diagnostic accuracy with only 1\% of training data, highlighting its exceptional learning ability and generalization performance in resource-limited environments.

\paragraph{Chest X-ray Segmentation}

Medical segmentation demands deep learning models to precisely delineate anatomical structures and identify pathological features in medical images, aiding in diagnosis. We focus on evaluating EVA-X's performance in both physiological and pathological segmentation tasks. Specifically, we fine-tune 7 different medical models~\cite{resnet, medklip, mgca, deit, biovil, xiao2023delving} across four lung segmentation tasks, encompassing physiological segmentation and pathological segmentation for pneumonia, pneumothorax, and tuberculosis. These tasks demonstrate the model's robust geometric understanding across diverse health conditions. Quantitative evaluation of segmentation results using Dice and Jaccard metrics, along with visualization of segmentation masks as depicted in Fig~\ref{fig:result2}, has been conducted through multiple experiments.

As shown in Fig~\ref{fig:result2} (a), EVA-X demonstrates outstanding performance across four distinct tasks~\cite{zhang2023lung, RSNA, siim, jaeger2014two}. Specifically, in lung segmentation, EVA-X achieves the highest average Dice score of 95.49\%. In pneumonia pathology segmentation, EVA-X surpasses both Medical MAE~\cite{xiao2023delving} (53.16 Dice, 36.20 Jaccard) and BioViL~\cite{biovil} (51.96 Dice, 35.10 Jaccard) with Dice and Jaccard scores of 54.51\% and 37.47\%, respectively. For pneumothorax pathology segmentation, EVA-X outperforms MGCA~\cite{mgca} (59.00 Dice, 41.84 Jaccard) and the ImageNet pretrained model~\cite{resnet} (57.69 Dice, 40.56 Jaccard) with scores of 60.27\% Dice and 43.13\% Jaccard. In pulmonary tuberculosis pathology segmentation, EVA-X excels with scores of 60.10\% Dice and 42.96\% Jaccard, surpassing Medical MAE~\cite{xiao2023delving} (59.1 Dice, 41.96 Jaccard) and MGCA~\cite{mgca} (59.00 Dice, 41.84 Jaccard). Furthermore, as illustrated in Fig~\ref{fig:result2} (b), EVA-X provides more accurate and fine-grained physiological or pathological segmentation, showcasing its exceptional generalization ability in X-ray segmentation tasks.

% Additionally, EVA-X is the most stable X-ray pre-trained model and has the smallest performance variance. For example, compared to the pre-trained X-ray model MGCA, EVA-X's performance stability improves by 78\%. In terms of visualization, EVA-X is able to give a more accurate and fine-grained physiological or pathological segmentation.

\subsection{Interpretability}

The interpretability of X-ray deep learning is an essential topic, as highlighted in Baselli et al.\cite{baselli2020opening}. Utilizing tools like the class activation map (CAM) can help elucidate the rationale behind neural network decisions, as discussed in Grad-CAM\cite{gradcam}. In medical domain, disease diagnosis often hinges on lesion localization. Saporta et al.~\cite{chexlocalize} have observed that while deep learning can provide reasonably accurate predictions, there remains a notable gap in its ability to automatically localize compared to human capabilities.

We employ Grad-CAM~\cite{gradcam} to analyze the gradients of EVA-X in the context of disease diagnosis. Our analysis involves approximately 1000 images from the Chest X-Ray14 dataset~\cite{cxr14}, as discussed in Sec.~\ref{sec:method-dataset} each annotated with lesion positions. Subsequently, we select seven different model weights pre-trained as outlined in Sec.~\ref{sec:pre-training} for comparative evaluation. We get Class Activation Maps (CAMs) with each pre-trained model and measure the Intersection over Union (IoU) and Average Precision (AP) between the activation regions and the ground truth (GT) boxes. To determine the optimal performance threshold, we conduct a search within the range of [0.1, 0.6].

We present the corresponding results in Figure~\ref{fig:result3} (a). Furthermore, we visually represent the CAM of EVA-X and the other six models using heatmaps, depicted in Figure~\ref{fig:result3} (b).
The results reveal several significant findings. Firstly, EVA-X demonstrates superior performance in terms of quantifiable metrics such as IoU, and AP compared to the other seven methods. Secondly, consistent with findings in prior research~\cite{xiao2023delving}, ViT pretrained with MAE exhibits notably weaker CAM performance than CNN. However, our experiments indicate a substantial enhancement in ViT's CAM quality when aided by EVA-X, resulting in a marked increase in mAP from 3.61 to 8.94. Additionally, our visual analysis highlights that EVA-X generates more accurate and distinct activation maps compared to previous methods. While CNN methods~\cite{biovil, mgca, medklip} exhibit superior map continuity, they may not perform as effectively as EVA-X in localizing smaller lesions.

\section{Discussion}

We propose EVA-X, medical foundation models tailored for X-ray images. Different from previous work, EVA-X utilizes a self-supervised pre-training strategy, combining previous pre-training methods of contrastive learning~\cite{mgca, biovil, medklip, convirt, gloria} and mask image modeling~\cite{selfmedmae, xiao2023delving}. It could learn generalizable visual representations for all X-ray tasks without any human-annotated images.  This unique advantage makes EVA-X's pre-training effective, efficient, flexible, and scalable. Compared with over 15 previous pre-trained deep learning models, EVA-X foundation models achieve state-of-the-art performance and computation trade-off. We transfer EVA-X models to 11 downstream tasks~\cite{cxr14, chexpert, hemdan2020covidx, zhang2023lung, siim, RSNA, jaeger2014two} and compare them with previous SOTA X-ray and natural image pre-trained models. The results show that EVA-X outperforms all previous models in all downstream tasks, demonstrating new state-of-the-art performance on X-ray image classification, segmentation, and interpretation. We argue EVA-X has great potential to become a general foundation model in medical X-ray analysis, facilitating faster and more accurate diagnosis and analysis of chest diseases.

\textbf{Limitations}
EVA-X are foundation models designed for medical X-ray images. Its training process is based entirely on data related to chest X-rays. Therefore, its performance on other medical tasks is open to improvement. Due to the unique high-performance self-supervised pre-training strategy of EVA-X and the great potential it shows on X-ray tasks, we believe that EVA-X's approach is expected to be extended to the entire medical field.

\section{Methods}
\label{sec:method}

\subsection{Dataset curation and pre-processing}
\label{sec:method-dataset}

\paragraph{Pre-training Dataset}
EVA-X is trained using exclusively public Chest X-Ray data. Our training set is a combination of three extensive public datasets: Chest X-Ray14~\cite{cxr14}, CheXpert~\cite{chexpert}, and MIMIC-CXR~\cite{mimic}. These datasets are widely recognized for their application in X-ray vision-language pre-training~\cite{mgca, medklip} and image classification~\cite{xprotonet, acpl}. In contrast to previous studies, our approach exclusively leverages pure unlabeled images for pre-training, without the utilization of any annotation or pathology report information.

For these datasets, we specifically process them as follows: (1) Following previous work~\cite{xiao2023delving}, we primarily use frontal view (AP/PA) images and discard lateral view images. (2) We do not use any of the images tested subsequently for training, even though they are unlabeled. (3) To speed up training, similar to CheXpert, we use bilinear interpolation to resize the original images to a resolution of 336\eq{\times}336. The combined dataset is called \textbf{Merged520k} (see Figure~\ref{fig:first-fig} (b)). If not otherwise noted, our pre-training experiments will be performed on this dataset.

\paragraph{Evaluation Dataset}
In the realm of natural images, ImageNet~\cite{imagenet} typically serves as the primary test dataset for pre-training~\cite{mae, eva, eva02}. Similarly, in the domain of X-ray images, it is essential to select a dataset for pre-training evaluation. Among the aforementioned datasets, both Chest X-Ray14 and CheXpert hold prominence as widely utilized categorization datasets~\cite{xprotonet, acpl, chexpert_sota}. They are characterized as multi-label categorical datasets, with labels assigned to 14 distinct diseases. Notably, these 14 labels are independent of each other. 

Here, we have opted for the former dataset, \textbf{Chest X-Ray14}, as our primary test set, which is the most commonly used X-ray classification dataset (as studied by~\cite{ccalli2021deep}). Our decision is based on the following reasons: (1) More rational dataset distribution. The CheXpert dataset comprises a total of 224k images, but only nearly 200 images are allocated for testing. In Chest X-Ray, the training/validation/test set ratio is 75k/11k/25k. (2) Clearer labeling. In the CheXpert dataset, the presence of an ``uncertain" annotation indicates that the physician did not identify the condition. Various approaches exist for handling this uncertainty. Some methods uniformly categorize it as "with disease," others as "without disease," and more complex treatment schemes are also employed. However, the labeling is clearer on the Chest X-Ray14 dataset. The selection of this test dataset is also consistent with the two previous works~\cite{selfmedmae, xiao2023delving}.

Note that this dataset selection indicates that we perform pre-training studies on this dataset, but does not mean that we only use this dataset to test the final performance of EVA-X. In subsequent sections, we will demonstrate the superior performance of EVA-X on additional datasets.

\subsection{EVA-X Architecture}

The pre-training process of EVA-X involves the design of the dual Vision Transformer (ViT)~\cite{vit} (see Figure~\ref{fig:enter-label}). The EVA-X transformer is learnable and the tokenizer is frozen. For the convenience of readers, we begin with a brief overview of ViT here.

Assuming the dimensions of the image are $H, W$, before attention calculation, ViT divides the image into $n = \frac{H}{P} \times \frac{W}{P}$ different patches, where $P$ represents the patch size. Typically, $P$ can take values like 16, 14, 8, etc. In EVA-X, unless specified otherwise, the patch size for all ViTs is set to 16. For an image patch, ViT uses linear projection to project it into a feature vector of dimension $d$, which is referred to as image tokens. These vectors form a sequence known as the image token sequence. Additionally, to establish positional relationships between vectors, ViT uses positional encoding added to the image token sequence. After adding the token dedicated to classification, we obtain the final input sequence, as shown in equation~\ref{eq1}, denoted as $Z$.
\begin{equation}
\label{eq1}
Z = \{z_0, z_1, \ldots, z_n\}
\end{equation}

The transformer block (see Figure~\ref{fig:enter-label} (b)) is a straightforward structure with the same output structure as the input. It mainly consists of two parts: Multi-Head Self-Attention (MHSA) and a Feed Forward Network (FFN). Inspired by Fang el. al.~\cite{eva02}, in EVA-X, we introduce improved structures such as rotational positional encoding, Sub-LN~\cite{subln}, and SwiGLU~\cite{swiglu}, which differ slightly from traditional ViT. By stacking any number of transformer blocks, the final ViT is composed. For the input $Z_i$ at layer $i$, the transformer block performs the following calculations to produce the final output $Z_{i+1}$, where $Z_i$ and $Z_{i+1}$ have the same structure.

\begin{equation}
Z_i' = \text{MHSA}(Z_i) + Z_i
\end{equation}
\begin{equation}
Z_{i+1} = \text{FFN}(Z_i') + Z_i'
\end{equation}

EVA-X is a learnable Vision Transformer. Here, we selected three ViTs of different sizes for experimentation: ViT-Ti, ViT-S, and ViT-B, with a patch size of 16 for each structure. Based on the number of parameters, we primarily use EVA-X-Ti (6M) to benchmark against DenseNet121~\cite{densenet} (8M), EVA-X-S (22M) against ResNet50~\cite{resnet}, and EVA-X-B (86M) to explore the effects and influences of scaling up the number of parameters.

To perform mask operations on images in mask image modeling, following previous work~\cite{eva, eva02}, we designed a mask token denoted as $m$. This token is a learnable $d$-dimensional vector. Assuming a mask ratio of $r$, we randomly replace $n \cdot r$ image tokens with mask tokens. We denote this sequence of masked tokens as \textit{mask\_list}. All mask tokens have the same initialization.

\begin{equation}
Z_{\text{e}} = \begin{cases} 
m & \text{if } i \in \text{mask\_list} \\
z_i & \text{otherwise}
\end{cases}
\end{equation}

Due to potential dimension differences between EVA-X and Tokenizer, we use a linear projection layer to map the dimension of EVA-X's image tokens from $d_{\text{eva\_x}}$ to $d_{\text{tgt}}$. We denote the final output sequence of EVA-X as
\begin{equation}
    Z_e = \{ze_0, ze_1, ze_2, \ldots, ze_n\}
\end{equation}

\subsection{Self-Supervised Learning}

The role of the Tokenizer is to extract semantically rich features from images, and it is also a ViT structure. Unlike EVA-X, we generally opt for larger-scale ViTs. We primarily investigate two types of structures for Tokenizer's pre-training performance, namely, natural image CLIP and medical image CLIP. For natural images, we selecte advanced high-performance ViT-B, ViT-L, and ViT-G visual encoders from the EVA-CLIP~\cite{eva_clip} model as our Tokenizer. In the medical field, we chose the ViT-B visual encoder trained with MGCA~\cite{mgca} as our Tokenizer. As far as we know, MGCA-ViT-B is currently the best open-source X-ray CLIP model. For more results of the influence of Tokenizer, please see our appendix.

Tokenizer takes the sequence \( Z \) as shown in the equation below as input and maps it to the target feature sequence \( Z_t \), represented by the following equation. During the pre-training process, all parameters of the Tokenizer are kept frozen, and no additional learnable linear mappings are added.
\begin{equation}
    Z_t = \{zt_0, zt_1, zt_2, \ldots, zt_n\}
\end{equation}

As mentioned earlier, for the token sequences in the equation \( Z \), we randomly select a proportion \( r \) of tokens and replace them with randomly initialized mask tokens. Here, we choose a relatively small mask ratio, \( r = 0.3 \). We denote the indices of the masked image tokens as \textit{mask\_list}.

For the final output sequences of EVA-X and Tokenizer, we respectively select the image tokens in \textit{mask\_list} to form the sequences \( Z_e' \) and \( Z_t' \). We aim to maximize the cosine similarity between corresponding tokens in \( Z_e' \) and \( Z_t' \), i.e.,
\begin{equation}
\text{maximize} \sum_{i=1}^{n \cdot r} \frac{Z_e'(i) \cdot Z_t'(i)}{\|Z_e'(i)\| \cdot \|Z_t'(i)\|}
\end{equation}

\subsection{Transfer Learning}

\paragraph{Classification}
In the case of classification tasks, we use the simplest decoding strategy uniformly for all models. For CNNs such as ResNet50~\cite{resnet}, DenseNet121~\cite{densenet}, etc., we average the features output from their last network layer for pooling, and then input the pooled features into a learnable linear layer to generate predictions. For the ViT~\cite{vit} structure used by methods such as EVA-X, we average all the tokens output from the last block, and then input the corresponding features into a learnable linear layer as well to output the prediction results. This simple structure ensures the ability to directly compare the underlying models, rather than a complex structural design.

We use the mean Area Under the Curve (mAUC) and mean Accuracy (mAcc) as our classification metric, as denoted in Eq~\ref{eq:auc} and \ref{eq:acc}, while \eq{TPR} denotes True Positive Ratio, \eq{FPR} denotes False Positive Ratio, \eq{TP} denotes True Positive, \eq{TN} denotes True Negative, \eq{FP} denotes False  Positive, and \eq{FN} denotes False Negative.

\begin{equation}
    \label{eq:auc}
    \text{AUC} = \int_{0}^{1} TPR(FPR) \, dFPR
\end{equation}

\begin{equation}
\label{eq:acc}
    \text{Accuracy} = \frac{TP + TN}{TP + TN + FP + FN}
\end{equation}

\paragraph{Segmentation}
Following the previous methods~\cite{medklip, mgca, biovil, gloria}, in this paper, we primarily focus on the comparison of pre-trained visual representation performance, without overly emphasizing the potential advantages that structural improvements may bring to the segmentation tasks. Specifically, we build two segmentation models using ResNet50~\cite{resnet} and ViT~\cite{vit} backbones, which are the most commonly used structures in X-ray pre-training. For ResNet50, we followed previous work~\cite{medklip, mgca}, adopting the structure with a ResNet encoder and a UNet~\cite{unet} decoder. For ViT, we follow common practices in natural images~\cite{vitdet}, initially building a feature pyramid by pooling and deconvolution on the last layer features, and then using UperNet~\cite{upernet} as the decoder for segmentation tasks. To ensure the simplicity of the structure as much as possible, we do not employ advanced adaptive structures, to better explore the performance of visual representations, although they may bring improvements in performance.

We use the mean of Dice and the mean of Jaccard as our segmentation metric, as shown in Eq~\ref{eq:dice} and Eq~\ref{eq:jaccard}, while \eq{S} denotes segmentation result and \eq{G} denotes ground truth.

\begin{equation}
    \label{eq:dice}
    \text{Dice} = \frac{2 \times |S \cap G|}{|S| + |G|}
\end{equation}

\begin{equation}
    \label{eq:jaccard}
    \text{Jaccard} = \frac{|S \cap G|}{|S \cup G|}
\end{equation}

\clearpage
\bibliography{sn-bibliography}% common bib file

\clearpage

\begin{appendices}

\section{Additional Results}

\subsection{Detail Classification Results}

\begin{figure}[h]
    \centering
    \includegraphics[width=\linewidth]{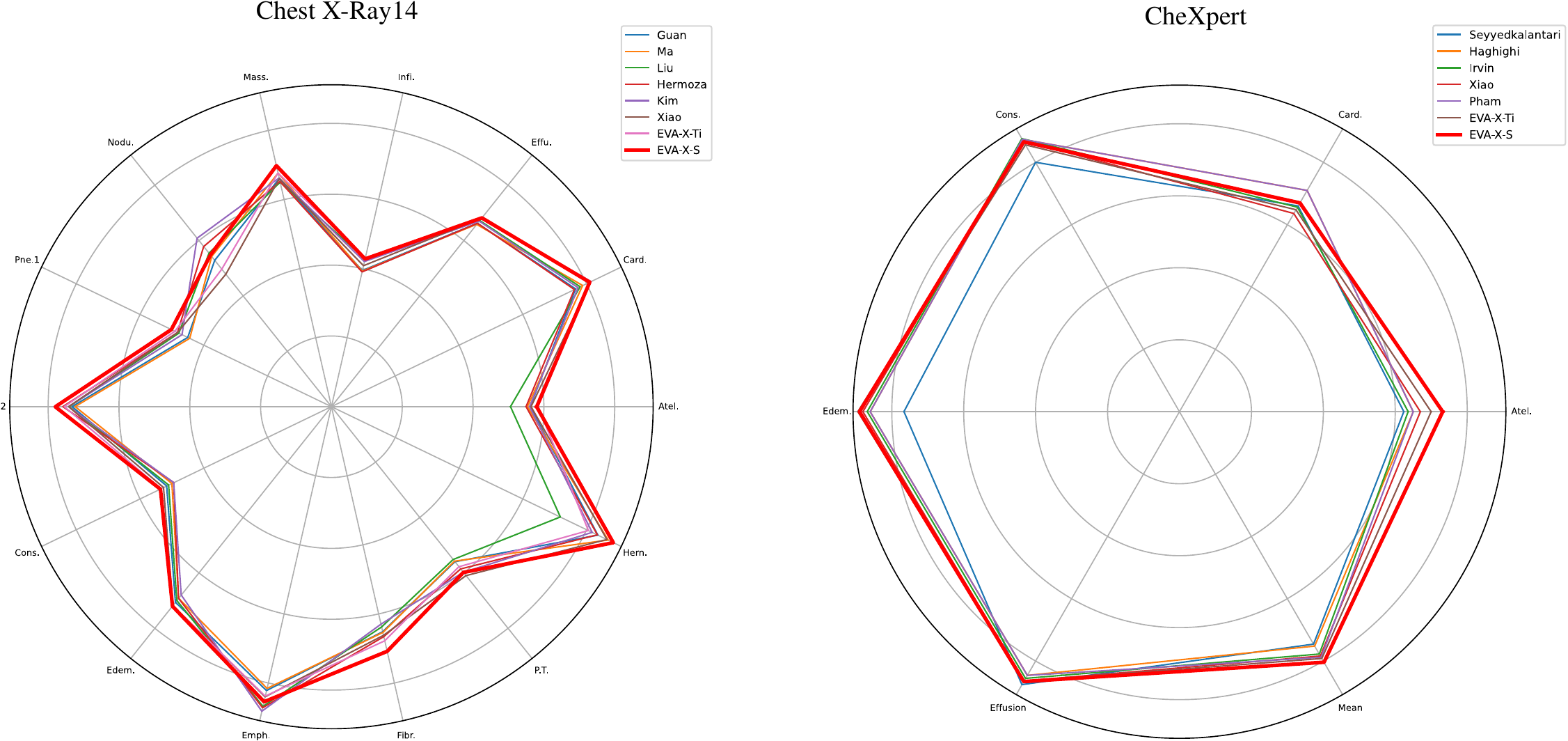}
    \caption{\textbf{Detail Results of Classification.} We compare EVA-X-Ti and EVA-X-S classification AUC with previous X-ray image classification tasks. EVA-X outperforms most of them and gets best performance in most diseases classification.}
    \label{fig:cls_detail}
\end{figure}

In the main mainuscript, we compare the average performance of EVA-X on a multi-disease classification task. Here, we provide a more detailed description of the classification results for both datasets.

As depicted in Figure~\ref{fig:cls_detail}, we present a comparative analysis of the classification results of EVA-X against various established methods on the multi-label classification benchmarks~\cite{cxr14, chexpert}. The outcomes for EVA-X-S are highlighted with a red line to emphasize its performance. Notably, EVA-X demonstrates superior performance across all diseases examined. Furthermore, within the Chest X-Ray14 dataset, the EVA-X series achieved the highest AUC scores in 12 out of 14 diseases. Similarly, on the CheXpert dataset, the EVA-X series outperformed competing methods in achieving the highest AUC scores in 2 out of 5 diseases.

\subsection{Pre-training Results}

\begin{table*}[tbp]
\subfloat[
\textbf{Tokenize EVA-X with CLIP(s)}. The CLIP series of visual models can all be employed as tokenizers for EVA-X, and their performance is consistently observed to surpass the reconstruction of pixel features. As the size of CLIP increases or its proprietary nature strengthens, the performance of EVA-X also improves accordingly.
\label{tab:reconstruct_targets}
]{
\begin{minipage}{0.92\linewidth}{\begin{center}
\tablestyle{10pt}{1.2}
        \begin{tabular}{lccc|c}
        \toprule
            \textbf{encoder} & \textbf{reconstruct targets} & \textbf{tokenizer} & \textbf{training epochs} & \textbf{mAUC} \\
        \midrule
            \multirow{5}{*}{ViT-S} & pixel desity values~\cite{xiao2023delving}                               & /     & 800 & 82.3   \\
        \cmidrule{2-5}
                                   & \multirow{2}{*}{medical CLIP features~\cite{mgca}}   & \multirow{2}{*}{ViT-B} & 300 & 82.7   \\
                                   
                                   & & & 600 & 83.3 \\ 
        \cmidrule{2-5}
                                   & \multirow{2}{*}{natural CLIP features~\cite{eva_clip}}   & ViT-B & \multirow{2}{*}{300}    & 82.2   \\
                                   &                                          & ViT-G &     & 82.7                       \\
        \bottomrule
        \end{tabular}
\end{center}}
\end{minipage}
}
\vspace{-.2em}
\subfloat[
\textbf{Encoder size}. Here, we train ViT-Ti for 900 epochs and ViT-S and ViT-B for 600 epochs. Performance of EVA-X improves with scaling of encoder size. 
\label{tab:encoder_size}
]{
\begin{minipage}{0.5\linewidth}{\begin{center}
\tablestyle{7pt}{1.05}
        \begin{tabular}{lcc|c}
        \toprule
            \textbf{encoder} & \textbf{\#params} &\textbf{tokenizer} & \textbf{mAUC} \\
        \midrule
            ViT-Ti  & 6M & \multirow{3}{*}{ViT-B}  & 82.4   \\
            ViT-S   & 22M &                        & 83.3   \\
            ViT-B   & 86M &                        & 83.5   \\
        \bottomrule
        \end{tabular}
\end{center}}
\end{minipage}
}
\hspace{0.5em}
\subfloat[
\textbf{Outperform the tokenizer}. The performance of EVA-X, pre-trained by mask image modeling, far exceeds the performance of Tokenizer. Even though the former model size is much smaller
\label{tab:outperform}
]{
\begin{minipage}{0.42\linewidth}{\begin{center}
\tablestyle{7pt}{1.05}
        \begin{tabular}{lc|c}
        \toprule
            \textbf{type} & \textbf{arch} & \textbf{mAUC} \\
        \midrule
            Tokenizer & ViT-B (MGCA) & 81.8       \\
            EVA-X     & ViT-S        & 83.3       \\
        \bottomrule
        \end{tabular}
\end{center}}
\end{minipage}
}
\vspace{-.2em}
\subfloat[
\textbf{Mask Ratio}. The performance impact of different mask ratios on downstream tasks. The experimental results show that the performance of the downstream task is optimized when the mask ratio is 0.3.
\label{tab:mask_ratio}
]{
\begin{minipage}{0.45\linewidth}{\begin{center}
\tablestyle{10pt}{1.05}
        \begin{tabular}{lc|c}
        \toprule
            \textbf{encoder} & \textbf{mask ratio} & \textbf{mAUC} \\
        \midrule
            \multirow{3}{*}{ViT-S} & 0.2 & 82.5     \\
                                   & 0.3 & 82.7     \\
                                   & 0.4 & 82.6     \\
        \bottomrule
        \end{tabular}
\end{center}}
\end{minipage}
}
\hspace{2em}
\subfloat[
\textbf{Crop Scale}. The effect of data-enhanced cropping ratios on training. We find that optimal pre-training performance is achieved using a randomized scale cropping of (0.2, 1).
\label{tab:crop_size}
]{
\begin{minipage}{0.45\linewidth}{\begin{center}
\tablestyle{10pt}{1.05}
        \begin{tabular}{lc|c}
        \toprule
            \textbf{encoder} & \textbf{crop scale} & \textbf{mAUC} \\
        \midrule
            \multirow{3}{*}{ViT-S} & 0.1 & 82.3       \\
                                   & 0.2 & 82.7       \\
                                   & 0.3 & 82.6       \\
        \bottomrule
        \end{tabular}
\end{center}}
\end{minipage}
}
\caption{Main experiments and ablations of EVA-X pre-trainings.}
\end{table*}

\paragraph{Tokenize X-ray Images with CLIP.} 

The reconstruction target of EVA-X has shifted from pixel density values to the feature vector of image tokens. A robust tokenizer is crucial for the training of EVA-X. We have uncovered the following insightful findings: 

As shown in Table~\ref{tab:reconstruct_targets}, Mask Image Modeling (MIM) with CLIP visual encoders consistently outperforms those with density values in pre-training. Recently, the use of CLIP visual encoder as a tokenizer for MIM has become a dominant trend in the field of natural images~\cite{eva, eva02, cae, caev2, mvp}. Inspired by them, we attempted to use natural CLIP~\cite{eva_clip}, which is widely trained on natural images, and medical CLIP~\cite{mgca}, which is specifically trained on X-ray images, as pre-trained tokenizers. We observe that both approaches demonstrated robust performance. Even with less than half the training length, our method showcases notably superior performance. For example, we have been able to achieve 82.7 mAUC by training EVA-X-S for only 300 epochs, surpassing the previous best result of 82.3 mAUC~\cite{xiao2023delving}, which was achieved by training for 800 epochs. Meanwhile, Further, when we train EVA-X-S for 600 epochs, EVA-X-S reaches 83.3 mAUC, a result that even exceeds the previous best pre-training result of ViT-B of 83.0 mAUC. 

Besides, we observe that both the proprietary nature and size enhancement of the tokenizer contribute to the improvement of EVA-X pre-training performance. In Table~\ref{tab:reconstruct_targets}, we use the natural CLIP ViT-B as the baseline. Increasing the tokenizer size from ViT-B to ViT-G results in EVA-X's performance rising from 82.2 to 82.7 on mAUC. Similarly, changing the weight of the ViT-B tokenizer from natural CLIP to medical CLIP enhances EVA-X's performance from 82.2 to 82.7. Unfortunately, to the best of our knowledge, there are no larger-sized X-Ray CLIPs available, and based on the above experiments, we suspect that EVA-X's performance will see further improvement when larger-sized X-Ray CLIPs become available.

\paragraph{Outperforming the Tokenizer.}
The toknenizer's performance cap does not limit the EVA-X's cap.
EVA-X demonstrates intriguing performance, where EVA, after pre-training with MIM, shows superior downstream task performance compared to its tokenizer teacher. 

As an example, the MGCA-trained ViT-B~\cite{mgca} achieves a direct finetune result of 81.8 mAUC (see Table~\ref{tab:outperform}) on the Chest X-Ray14 dataset. When we use ViT-B (MGCA) as a tokenizer for EVA-X, the results of our trained EVAs are presented in Table~\ref{tab:encoder_size}. The ViT series obtained by EVA-X outperforms the original tokenizer across all sizes. The ViT-B (EVA-X) outperforms the ViT-B (MGCA) by 1.7 mAUC with the same number of parameters, and even the ViT-Ti (EVA-X), with only 6M parameters, outperforms the ViT-B (MGCA) by 0.4 mAUC. This MIM pre-trained model surpasses the observations of the original tokenizer and aligns with findings from pre-training tasks with natural images~\cite{maskfeat, beitv2}. This suggests that the performance of the tokenizer does not dictate the upper limit of the MIM pre-training task. We find this encouraging as it provides an opportunity to enhance the original model with unsupervised data and MIM, as demonstrated by EVA-X.

\textbf{Scalability of Model Size.} 
The EVA-X model demonstrates superior parameter scaling performance. Considering the number of models and parameters used in previous X-Ray methods, such as DenseNet121 (6M)~\cite{densenet} and ResNet50(26M)~\cite{resnet}, we have pre-trained three different EVA-X models: ViT-Ti (6M), ViT-S (22M), and ViT-B (86M). As depicted in Table~\ref{tab:encoder_size}, increasing the number of parameters consistently results in performance improvement. When the total number of parameters is small, augmenting the model parameters yields more substantial improvement; however, as the total number of parameters reaches a certain threshold, the performance improvement derived from increasing the number of parameters becomes less significant.

\begin{figure}
    \centering
    \includegraphics[width=\linewidth]{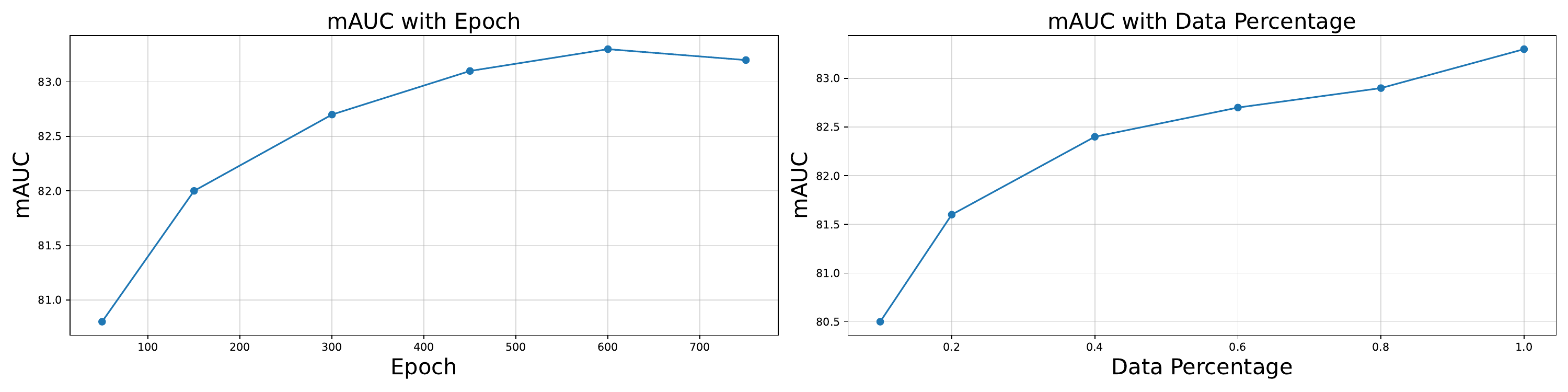}
    \caption{\textbf{Fast Convergence (left)},  EVA-X has faster convergence than previous methods~\cite{selfmedmae, xiao2023delving}. \textbf{Scalability with Pre-training Data (right).} The performance of EVA-X improves with pre-training data.}
    \label{fig:scale}
\end{figure}

\textbf{Scalability of Pre-training Data.} 
In previous studies, the ViT-S model appeared to manifest data saturation issues~\cite{xiao2023delving}. Specifically, when RGB values are used as the reconstruction target, it is observed that the performance of ViT-S reached saturation at around 300k data, and further increases in data no longer led to performance improvement. This is a frustrating phenomenon, as the primary advantage of unsupervised pretraining lies in the ability to leverage an infinite amount of data.

However, there appears to be a positive shift here at EVA-X. As illustrated in the Figure~\ref{fig:scale} right, we randomly divide Merged520k into 10\%, 20\%, 40\%, 60\%, 80\%, and full-volume data. We observe that as the data volume increases, the downstream task performance of EVA-X improves further. This suggests that the pre-training of EVA-X can effectively utilize more data and shows greater data scalability. We attribute this enhancement to a shift in the training strategy of EVA-X, transitioning from reconstructing RGB values to reconstructing semantic information.

\begin{figure}
    \centering
    \includegraphics[width=\linewidth]{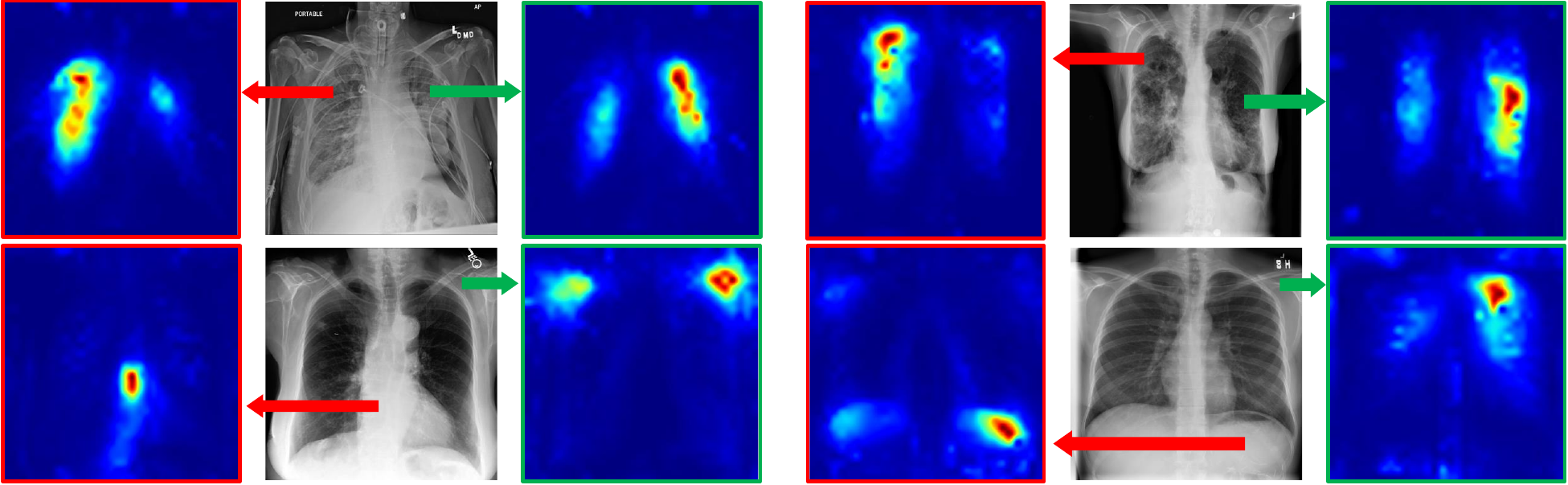}
    \caption{\textbf{Visualization of EVA-X Self-attention.} After pre-training EVA-X automatically produces some segmentation results with actual physical meaning, even though we did not use any labeling during the training process.}
    \label{fig:attention_map}
\end{figure}

\paragraph{Attention Map Visualization.} 
We present the visualization of self-attention in the ViT-S model after EVA-X pre-training. In the attention computation process, each image token is projected as a query, key, and value (Q, K, V). The attention mechanism signifies the degree of response of each query to all keys. For a given input image, we designate specific reference points and visualize the heatmap corresponding to its query's attention map.

As observed in the Figure~\ref{fig:attention_map}, the self-attention of EVA-X after pre-training demonstrates notable semantic perception ability. It autonomously segments regions with tangible anatomical meaning, such as the lungs, scapula, spine, etc. Moreover, the attention map also indicates that EVA-X has learned the human body's symmetry. For instance, even though our reference point is only in the right lung, the degree of attention response corresponding to the left lung is remarkably high, forming a double-lung segmentation map. This pre-training performance, which enables automatic segmentation of images, is similar to what has been observed in well-known work BEiT~\cite{beit}, DINO~\cite{dino}, and DINOv2~\cite{dinov2} on the pre-training of natural images.

\textbf{Faster convergence.} Compared to previous methods, EVA-X has a faster convergence rate. As shown in Figure~\ref{fig:scale} left, we use ViT-B MGCA as a tokenizer and train EVA-X-S for 50, 150, 300, 450, 600, and 750 epochs respectively. We find that the performance of EVA-X-S converges at about 600 epochs of training. This convergence rate outperforms the previous 800-epoch convergence methods~\cite{selfmedmae, xiao2023delving}.

\textbf{Hyper-parameters.} We focus on two important parameters, mask ratio, and crop ratio, during pre-training. The setting of these two parameters of EVA-X is mainly derived from experimental findings. We compare mask ratio settings of 0.2, 0.3, and 0.4, and find that a mask ratio of 0.3 gave the best performance in EVA-X (see Table~\ref{tab:mask_ratio}). This ratio is slightly smaller than the natural image. For data enhancement, we set the minimum cropping ratios to 0.1, 0.2, and 0.3. we found that the performance of EVA-X was optimized when using 0.2. This cropping ratio is smaller than previous work~\cite{xiao2023delving}.

\section{Additional Experiment Details}

Here we provide implementation details of EVA-X. One could reproduce all of our experiments in the paper based on our codes here: \url{https://github.com/hustvl/EVA-X}.

\subsection{Pre-training Details}

Details of the specific pre-training for EVA-X are as follows. EVA-X is available in three different sizes, each embedding an image with a patch size of 16, namely, EVA-X-Ti/16, EVA-X-S/16, and EVA-X-B/16. Their corresponding parameter counts are 6M, 22M, and 86M, respectively. We use memory-efficient attention provided by \textit{xformers}~\cite{xformers} to expedite training. For the tokenizer, we use the ViT-B visual encoder from the pre-trained vision-language model of MGCA~\cite{mgca} and ViT-B/L/G visual encoders from advanced natural CLIP models, EVA-CLIP~\cite{eva_clip}. Their parameters remain entirely frozen during training. We perform simple data augmentation on the pre-trained data. This includes random horizontal flipping, random scaling, and fixed-size cropping. The random scaling range is [3/4, 4/3], and the fixed cropping size is 224. Subsequently, we normalize all input images using the mean and variance of Merged520k. We use AdamW with \eq{\beta_1=0.9, \beta_2=0.98} as our optimizer. The learning rate is set at 3e-4, and a cosine annealing descent strategy is employed. During the pre-training process, we use 8 A100 GPUs and train EVA-X for 600 epochs to obtain the final results of EVA-X-S and EVA-X-B, while 900 epochs for EVA-X-Ti.

\subsection{Classification Transfer Learning Details}

\subsubsection{Classification Dataset}

\textbf{Chest X-Ray14} is proposed by \mycitename{Wang}~\cite{cxr14}. As mentioned above, due to its high-quality data and extensive research and use, we use it as a standard test set for pre-training. It contains a total of 111,120 chest x-ray images of frontal view, each of which measures 1024 and is labeled by a medical professional with 14 different disease infections such as Atelectasis, Consolidation, and others. Note that while these diseases may be correlated in a medical sense, in our actual experiments we consider these 14 different diseases to be independent of each other. Following \mycitename{Xiao}~\cite{xiao2023delving}, each of the images is resized to 224 while training and testing.

\textbf{CheXpert} is proposed by \mycitename{Irvin}~\cite{chexpert}. Its training and validation sets are publicly available, but the test set is not. It contains a total of 224,316 publicly available frontal view or lateral view Chest X-ray images, of which the number of test set images is only 236. There is a certain gap between its training set and validation set.First, its training set contains 14 classes of labeling while the validation set has only 5 classes of labeling. Second, in the training set, for each class of disease, there is a labeling marked as -1 in addition to the two labelings 0,1. When this type of disease is labeled as -1, it means that the presence or absence of the disease is doubtful and has not been fully determined by the doctor. In the validation set, this is not the case. Following \mycitename{Pham}~\cite{pham2021interpreting}, each of the images is resized to 224 while training and testing.

\textbf{CovidX} is originally proposed by \mycitename{Wang}~\cite{Wang2020}. In this paper, we use version of \textbf{CovidX-CXR-3 Version5} and \textbf{CovidX-CXR-4 Version9} \footnote{Different versions of CovidX can be obtained on \url{https://www.kaggle.com/datasets/andyczhao/covidx-cxr2}}. The former contains about 30k training images and the latter contains about 67k training images. Both datasets focus on the detection effect of the disease COVID-19.
Following previous work~\cite{Wang2020}, we used Covid-19 sensitivity and mean of accuracy to evaluate the final metrics. Each of the images is resized to 224 while training and testing.

\subsubsection{Implementation Details}

Our model is deterministic for different datasets. Specifically, we use the EVA-X-Ti and EVA-X-S trained in the previous section of pre-training to output the final result by a linear projection. We use mean-pooling of the feature map instead of the class token since we observe a slight improvement in fine-tuning. For two different tasks, we train using BCE loss and CE loss, respectively. For training, we use fixed learning rate optimization with AdamW and 1e-3. Layer-wise Learning Rate Decay (LLRD) is also used, with the parameter set to 0.55. drop-out ratio is set to 0.2. Each task is trained on 4 RTX 3090 GPUs. Batch size per GPU was set to 256 or 128. For Chest X-Ray14, we train with it for 30 epochs, and for the last 3 datasets, we train with them for only 10 epochs.

\subsection{Segmentation Transfer Learning Details}

\subsubsection{Segmentation Dataset}

\textbf{Lung Segmentation~\cite{zhang2023lung}.} This dataset is a recent lung segmentation dataset proposed by \mycitename{Zhang}~\cite{zhang2023lung}. They select 1000 images from the RSNA Pneumonia Detection~\cite{RSNA} dataset and invite professional doctors to annotate the lungs. The annotated images are divided into 800 training images and 200 testing images. In our experiments, all training and testing images are resized to a resolution of 512.

\textbf{SIIM-ACR Pneumothorax Segmentation~\cite{siim}} is a well-known Kaggle challenge. The dataset is provided by \mycitename{Zawacki}. It is divided into two stages, and here we use the dataset from stage 2. It includes 3205 training images, of which 2669 images are labeled for pneumonia, and corresponding masks are provided. Following previous work~\cite{mgca}, we randomly split this dataset into 1868 training images and 801 testing images.

\textbf{RSNA Pneumonia Detection} is also a Kaggle challenge provided by \mycitename{Stein}~\cite{RSNA}. We use the dataset from its second stage. The dataset comprises 26.7k training images and 3000 testing images. Among them, 6012 images are labeled for pneumonia, and corresponding location information for annotations is provided. Following the same processing approach as in previous work, we divide it into 4236 training images and 1776 testing images.

\textbf{Shenzhen Hospital Chest X-ray.~\cite{jaeger2014two}} This dataset comprises 662 chest X-ray images, with 326 being normal and 336 being abnormal. For these 336 abnormal images, the dataset provides a binary mask annotation to locate the appearance of pulmonary tuberculosis symptoms. In our experiments, we randomly split these 336 images into 236 training images and 100 testing images. Throughout the training and testing processes, all images are resized to a resolution of 512.

\subsubsection{Implementation Details}

For a fair comparison, all experiments are conducted based on the mmsegmentation~\cite{mmseg2020} framework. We construct our convolutional models using commonly used public libraries~\cite{Iakubovskii:2019}. For the ResNet method~\cite{resnet}, we use the SGD optimizer with a learning rate set at 0.01, and weight decay set to 0.0005. For the ViT method~\cite{vit}, we use the AdamW optimizer with a learning rate set at 2e-4. Layer-wise learning rate decay is set to 0.85. All models are trained on a single RTX 3090 GPU, with a batch size set to 4. For the aforementioned datasets, we train for either 4k or 10k iterations. The hyperparameters for different visual representations of the same structure are identical, with only differences in the initialization method. During training, we apply data augmentation operations such as resizing and random flipping, and all training and testing data are processed at a resolution of 512.

\end{appendices}

\end{document}